\documentclass[11pt,letterpaper]{article}
\usepackage{naaclhlt2015}
\usepackage{times}
\usepackage{latexsym}
\usepackage{amsmath}
\usepackage{graphicx}
\usepackage{multirow}
\usepackage{colortbl}
\usepackage{graphicx}
\usepackage{caption}
\usepackage{subcaption}
\usepackage{color}
\usepackage{url}
\usepackage{tikz,pgfplots}
\usepackage [english]{babel}
\usepackage [autostyle, english = american]{csquotes}
\MakeOuterQuote{"}

\title{And That's A Fact: \\ Distinguishing Factual and Emotional
  Argumentation in Online Dialogue}

\author{}
   \author{
   \textbf{Shereen Oraby$^*$, Lena Reed$^*$, Ryan Compton$^*$,} \\
   \textbf{Ellen Riloff $^\dag$, Marilyn Walker$^*$ and Steve Whittaker$^*$} \\
   $^*$ University of California Santa Cruz \\ {\tt \{soraby,lireed,rcompton,mawalker,swhittak\}@ucsc.edu} \\
   $^\dag$ University of Utah \\ {\tt riloff@cs.utah.edu}  \\ 
   }
\date{}

\begin{document}
\maketitle
\begin{abstract}

  We investigate the characteristics of factual and emotional
  argumentation styles observed in online debates. Using an annotated
  set of {\sc factual} and {\sc feeling} debate forum posts, we
  extract patterns that are highly correlated with factual and
  emotional arguments, and then apply a bootstrapping methodology to
  find new patterns in a larger pool of unannotated forum posts.  This
  process automatically produces a large set of patterns representing
  linguistic expressions that are highly correlated with factual and
  emotional language. 
  Finally, we analyze the most discriminating
  patterns to better understand the defining characteristics of 
    factual and emotional arguments.

\end{abstract}

\section{Introduction}
\label{sec:intro}

Human lives are being lived online in transformative ways: people can
now ask questions, solve problems, share opinions, or discuss current
events with anyone they want, at any time, in any location, on any
topic. The purposes of these exchanges are varied, but a significant
fraction of them are argumentative, ranging from hot-button political
controversies (e.g., national health care) to religious interpretation
(e.g., Biblical exegesis). And while the study of the structure of
arguments has a long lineage in psychology \cite{Cialdini00} and
rhetoric \cite{Hunter87}, large shared corpora of natural informal
argumentative dialogues have only recently become available. 

Natural informal dialogues exhibit a much broader range of
argumentative styles than found in traditional work on argumentation
\cite{Marwell67,Cialdini00,mcalisteretal14,ReedRowe04}.  Recent work
has begun to model different aspects of these natural informal
arguments, with tasks including stance classification
\cite{SomasundaranWiebe10,Walkeretal12c}, argument summarization
\cite{Misraetal15}, sarcasm detection \cite{Justoetal14}, and work on
the detailed structure of arguments
\cite{BiranRambow11,Purpuraetal08,YangCardie13}.  Successful models of
these tasks have many possible applications in sentiment detection,
automatic summarization, argumentative agents \cite{Zuckermanetal15}, 
and in systems that support human argumentative behavior
\cite{RosenfeldKraus15}.

Our research examines {\sc factual} versus {\sc feeling} argument
styles, drawing on annotations provided in the 
Internet Argument Corpus (IAC) \cite{Walkeretal12c}.
This corpus includes
quote-response pairs that were manually annotated with respect to
whether the response is primarily a {\sc factual} or {\sc feeling}
based argument, as Section~\ref{data-sec} describes in more detail.
Figure~\ref{sample-fact-feeling} provides examples of responses in the
IAC (paired with preceding quotes to provide context), along with the
response's {\sc factual} vs. {\sc feeling} label.

{\sc factual} responses may try to bolster their argument by providing
statistics related to a position, giving historical or scientific
background, or presenting specific examples or data.  There is clearly
a relationship between a proposition being {\sc factual} versus {\sc
  objective} or {\sc veridical}, although each of these different
labelling tasks may elicit differences from annotators
\cite{wiebe05,riloff-emnlp03,SauriPustejovsky09,ParkCardie14}.

\begin{figure}[!thb]
\begin{center}
\begin{small}
\begin{tabular}{|p{.35in}|p{2.5in}|}
\hline  
Class & Debate Forum Dialogue \\ \hline
{\sc FACT} & 
{\bf Quote:}  Even though our planet is getting warmer, it is still a lot cooler than it was 4000 years ago. \\
 & {\bf Response:}  \textit{The average global temperature follows a sinusoidal pattern, 
 the general consensus is we are supposed to be approaching a peak. 
 Projections show that instead of peaking, there will be continue to be an 
 increase in average global temperature.}
\\ \hline  \hline
{\sc FACT} & 
{\bf Quote:} 
"When you go to war against your enemies...suppose you see a 
beautiful woman whom you desire...you shall take her..and she 
shall marry you." - Deut. 21:10 \\
& {\bf Response:}
\textit{Read to the very end of the verse. "If you are not pleased 
with her, let her go wherever she wishes. You must not sell 
her or treat her as a slave, since you have dishonored her."}
\\ \hline  \hline
{\sc FEEL} & 
{\bf Quote:} Talk about begging the question! I don't want your gun, 
and if such a law were passed it's not my job to enforce the law. \\ 
& {\bf Response:} \textit{I see you are willing to violate my constitutional rights yet you expect 
someone else to do your dirty work.... How typical.}
\\ \hline  \hline
{\sc FEEL} & 
{\bf Quote:} 
"WASHINGTON \&\#8211; Supreme Court aspirant Sonia Sotomayor 
said Tuesday that she considers the question of abortion 
rights is settled precedent and says there is a constitutional 
right to privacy. The federal appeals court judge was asked at 
her confirmation...." \\
& {\bf Response:}
\textit{While I'm still iffy on her with the whole New Haven case, 
and her off-the-bench comments on race, this is one thing 
I commend her for and agree completely with.}
\\ \hline  
\end{tabular}
\end{small}
\end{center}
\caption{\label{sample-fact-feeling} Examples of {\sc factual} and
  {\sc feeling} based debate forum Quotes and Responses. Only the responses
were labeled for {\sc fact} vs. {\sc feel}.}
\end{figure}

The  {\sc feeling} responses may seem to lack
argumentative merit, but previous work on argumentation describes
situations in which such arguments can be effective,
such as the use of emotive arguments to draw attention away from
the facts, or to frame a discussion in a particular way
\cite{Walton10,MacagnoWalton14}. Furthermore, work on 
persuasion suggest that {\sc feeling} based arguments can be more
persuasive in particular circumstances, such as when the hearer shares
a basis for social identity with the source (speaker)
\cite{Chaiken80,PettyCacioppo86,Benoit87,Cacioppoetal83,Pettyetal81}.
However none of this work has documented the linguistic patterns that
characterize the differences in these argument types, which is a
necessary first step to their automatic recognition or classification.
Thus the goal of this paper is to use computational methods for
pattern-learning on conversational arguments to catalog linguistic
expressions and stylistic properties that distinguish Factual from
Emotional arguments in these on-line debate forums.

Section~\ref{data-sec} describes the manual annotations for
{\sc factual} and {\sc feeling} in the IAC corpus.  Section
\ref{sys-method} then describes how we generate lexico-syntactic
patterns that occur in both types of argument styles.  We use a weakly
supervised pattern learner in a bootstrapping framework to
automatically generate lexico-syntactic patterns from both annotated
and unannotated debate posts.  
Section \ref{results-sec} evaluates the precision and recall of the
{\sc factual} and {\sc feeling} patterns learned from the annotated
texts and after bootstrapping on the unannotated
texts. We also present results for a supervised learner with bag-of-word
features to assess the difficulty of this task. 
Finally, Section \ref{analysis-sec} presents analyses of
the linguistic expressions found by the pattern learner and presents
several observations about the different types of linguistic
structures found in {\sc factual} and {\sc feeling} based argument
styles.  Section \ref{related-sec} discusses related research,
and Section~\ref{conc-sec} sums up and proposes possible avenues for future work.

\section{Pattern Learning for Factual and Emotional Arguments}
\label{method-sec}

We first describe the corpus of online debate posts
used for our research, and then present a bootstrapping method to
identify linguistic expressions associated with {\sc factual} and {\sc
  feeling} arguments.

\subsection{Data}
\label{data-sec}

The IAC corpus is a freely available annotated collection of 109,553
forum posts (11,216 discussion threads). \footnote{{\tt
    https://nlds.soe.ucsc.edu/iac }} 
In such forums, conversations are started 
by posting a topic or a question in a particular
category, such as society, politics, or religion \cite{Walkeretal12c}.
Forum participants can then post their opinions, choosing whether to
respond directly to a previous post or to the top level topic (start a
new thread).  These discussions are essentially dialogic; however the
affordances of the forum such as asynchrony, and the ability to start
a new thread rather than continue an existing one, leads to dialogic
structures that are different than other multiparty informal
conversations \cite{FoxTree10}. An additional source of dialogic
structure in these discussions, above and beyond the thread structure,
is the use of the quote mechanism, which is an interface feature that
allows participants to optionally break down a previous post into the
components of its argument and respond to each component in turn.

The IAC includes 10,003 Quote-Response (Q-R) pairs with annotations for {\sc factual}
vs. {\sc feeling} argument style, across a range of
topics. Figure~\ref{mturk-survey-fig} shows the wording of the survey
question used to collect the annotations. Fact vs. Feeling was
measured as a scalar ranging from -5 to +5, because previous work
suggested that taking the means of scalar annotations reduces noise in
Mechanical Turk annotations \cite{Snowetal08}. Each of the pairs was annotated
by 5-7 annotators.  

For our experiments, we use only the response texts and assign a
binary {\sc Fact} or {\sc Feel} label to each response: texts with 
score $>$ 1 are assigned to the {\sc fact} class and texts with score $<$ -1
are assigned to the {\sc feeling} class.  We did not use the responses
with scores between -1 and 1 because they had a very weak Fact/Feeling
assessment, which could be attributed to responses either containing aspects of \textit{both} factual and feeling expression, or neither. The resulting set contains 3,466 {\sc fact}
and 2,382 {\sc feeling} posts. We randomly partitioned the {\sc fact}/{\sc feel} responses into
three subsets: a training set with 70\% of the data (2,426 {\sc fact}
and 1,667 {\sc feeling} posts), a development (tuning) set with 20\%
of the data (693 {\sc fact} and 476 {\sc feeling} posts), and a test
set with 10\% of the data (347 {\sc fact} and 239 {\sc feeling}
posts).  For the bootstrapping method, we also used 11,560 responses
from the unannotated data.



\begin{figure}[!htb]
\begin{center}
\begin{small}
\begin{tabular}{|p{3.0in}|}
\hline  
Slider Scale -5,5: Survey Question \\ \hline  \hline  
{\bf Fact/Emotion}: Is the respondent attempting to make a fact based argument or appealing to feelings and emotions?\\ \hline  
\end{tabular}
\end{small}
\end{center}
\vspace{-.1in}
\caption{\label{mturk-survey-fig} Mechanical Turk Survey Question used for Fact/Feeling annotation.}
\vspace{-.2in}
\end{figure}

\subsection{Bootstrapped Pattern Learning}
\label{sys-method}

The goal of our research is to gain insights into the types of
linguistic expressions and properties that are distinctive and common in factual and
feeling based argumentation. We also explore whether it is possible to 
develop a high-precision {\sc fact} vs. {\sc feeling} classifier that can
be applied to unannotated data to find new linguistic
expressions that did not occur in our original labeled corpus. 

To accomplish this, we use the AutoSlog-TS system \cite{riloff1996} to extract linguistic
expressions from the annotated texts. Since the IAC also contains a
large collection of unannotated texts, we then embed
AutoSlog-TS in a bootstrapping framework to learn additional
linguistic expressions from the unannotated texts. First, we briefly
describe the AutoSlog-TS pattern learner and the set of pattern templates
that we used. Then, we present the bootstrapping
process to learn more Fact/Feeling patterns from unannotated texts. 

\subsubsection{Pattern Learning with AutoSlog-TS}

To learn patterns from texts labeled as {\sc fact} or {\sc feeling}
arguments, we use the AutoSlog-TS \cite{riloff1996} extraction
pattern learner, which is freely available for research.  AutoSlog-TS
is a weakly supervised pattern learner that requires training data consisting
of documents that have been labeled with respect to different
categories. For our purposes, we provide AutoSlog-TS with 
responses that have been labeled as either {\sc fact} or {\sc feeling}. 

AutoSlog-TS uses a set of syntactic templates to define different
types of linguistic expressions. The left-hand side of Figure
\ref{pattern-types} shows the set of syntactic templates defined 
in the AutoSlog-TS software package.  PassVP refers to passive voice verb
phrases (VPs), ActVP refers to active voice VPs, InfVP refers to
infinitive VPs, and AuxVP refers to VPs where the main verb is a form
of ``to be'' or ``to have''. Subjects (subj), direct objects (dobj),
noun phrases (np), and possessives (genitives) can be extracted by the patterns.
AutoSlog-TS applies the Sundance shallow parser
\cite{RiloffPhillips04} to each sentence and finds every possible
match for each pattern template. For each match, the template
is instantiated with the corresponding words in the sentence to
produce a specific lexico-syntactic expression. The right-hand side of
Figure \ref{pattern-types} shows an example of a specific
lexico-syntactic pattern that corresponds to each general pattern
template.\footnote{The examples are shown as general expressions for
  readability, but the actual patterns must match the syntactic
  constraints associated with the pattern template.}

\begin{figure}[ht]
  \small
  \centering
  \begin{tabular}{|ll|}    \hline
    {\sc {\bf Pattern Template}} & {\sc {\bf Example Pattern}} \\
    \hline
    $<$subj$>$ PassVP & $<$subj$>$ was observed \\
    $<$subj$>$ ActVP        & $<$subj$>$ observed \\
    $<$subj$>$ ActVP Dobj   & $<$subj$>$ want explanation \\
    $<$subj$>$ ActInfVP & $<$subj$>$ expected to find \\
    $<$subj$>$ PassInfVP & $<$subj$>$ was used to measure \\
    $<$subj$>$ AuxVP Dobj & $<$subj$>$ was success \\ 
    $<$subj$>$ AuxVP Adj & $<$subj$>$ is religious \\
    \hline
    ActVP $<$dobj$>$ & create $<$dobj$>$ \\
    InfVP $<$dobj$>$ & to limit $<$dobj$>$ \\
    ActInfVP $<$dobj$>$ & like to see $<$dobj$>$ \\ 
    PassInfVP $<$dobj$>$ & was interested to see $<$dobj$>$ \\ 
    Subj AuxVP $<$dobj$>$   & question is $<$dobj$>$ \\
    \hline
    NP Prep $<$np$>$        & origins of $<$np$>$ \\
    ActVP Prep $<$np$>$     & evolved from $<$np$>$ \\
    PassVP Prep $<$np$>$ & was replaced by $<$np$>$ \\  
    InfVP Prep $<$np$>$ & to use as $<$np$>$ \\ 
    $<$possessive$>$ NP & $<$possessive$>$ son \\    \hline
  \end{tabular}
  \caption{The Pattern Templates of AutoSlog-TS with Example Instantiations}
  \label{pattern-types}
\end{figure}

In addition to the original 17 pattern templates in AutoSlog-TS (shown
in Figure \ref{pattern-types}), we defined 7 new pattern templates for
the following bigrams and trigrams: {\tt Adj Noun}, {\tt Adj Conj
  Adj}, {\tt Adv Adv}, {\tt Adv Adv Adv}, {\tt Adj Adj}, {\tt Adv
  Adj}, {\tt Adv Adv Adj}.  We added these n-gram patterns to provide
coverage for adjective and adverb expressions because the original
templates were primarily designed to capture noun phrase and verb
phrase expressions.

The learning process in AutoSlog-TS has two phases. In the first phase, the
pattern templates are applied to the texts exhaustively, so that
lexico-syntactic patterns are generated for (literally) every instantiation of the
templates that appear in the corpus.  In the second phase,
AutoSlog-TS uses the labels associated with the texts to compute
statistics for how often each pattern occurs in each class of texts. 
For each pattern $p$, we collect P({\sc factual} $\mid$ $p$) and
P({\sc feeling} $\mid$ $p$), as well as the pattern's overall frequency in the
corpus. 

\subsubsection{Bootstrapping Procedure}
\label{bootstrap-section}

\begin{figure}[h]
  \centering
    \includegraphics[width=0.95\columnwidth]{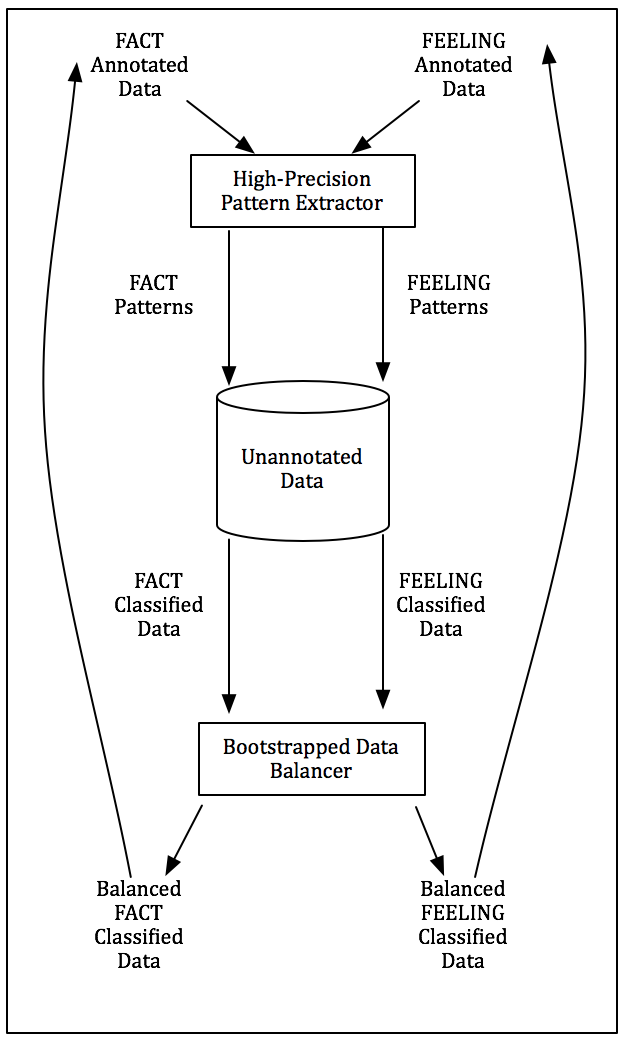}
  \caption{Flow Diagram for Bootstrapping Process}
  \label{fig:system}
\end{figure}

Since the IAC data set contains a large number of unannotated debate
forum posts, we embedd AutoSlog-TS in a bootstrapping framework to
learn additional patterns. The flow diagram for the 
bootstrapping system is shown in Figure \ref{fig:system}.

Initially, we give the labeled training data to AutoSlog-TS, which
generates patterns and associated statistics. The next step identifies
high-precision patterns that can be used to label some of the
unannotated texts as {\sc factual} or {\sc feeling}.  We define two
thresholds: $\theta_{f}$ to represent a minimum frequency value, and
$\theta_{p}$ to represent a minimum probability value.  We found that using
only a small set of patterns (when $\theta_{p}$ is set to a high value) achieves extremely high precision, yet results in a very low recall. Instead, we adopt a strategy of setting a moderate probability threshold to identify
reasonably reliable patterns, but labeling a text as {\sc factual} or
{\sc feeling} only if it contains at least a certain number different patterns for
that category, $\theta_{n}$. In order to calibrate the thresholds, we experimented with a range of
threshold values on the development (tuning) data and identified
$\theta_{f}$=3, $\theta_p$=.70, and $\theta_{n}$=3 for the {\sc factual} class, and
$\theta_{f}$=3, $\theta_p$=.55, and $\theta_{n}$=3  for the {\sc feeling} class as having the highest
classification precision (with non-trivial recall).


The high-precision patterns are then used in the bootstrapping
framework to identify more {\sc factual} and {\sc feeling} texts from
the 11,561 unannotated posts, also from 4forums.com. For each round of
bootstrapping, the current set of {\sc factual} and {\sc feeling}
patterns are matched against the unannotated texts, and posts that
match at least 3 patterns associated with a given class are assigned
to that class.
As shown in Figure \ref{fig:system}, the Bootstrapped Data Balancer then
randomly selects a balanced subset of the newly classified posts to
maintain the same proportion of {\sc factual} vs. {\sc feeling} documents
throughout the bootstrapping process. These new documents are added to
the set of labeled documents, and the bootstrapping process repeats.
We use the same threshold values to select new high-precision patterns
for all iterations. 

\section{Evaluation}
\label{results-sec}

We evaluate the effectiveness of the learned patterns by applying them
to the test set of 586 posts (347 {\sc fact} and 239 {\sc feeling}
posts, maintaining the original ratio of {\sc fact} to {\sc feel} data in train). 
We classify each post as {\sc factual} or {\sc feeling} using the same
procedure as during bootstrapping: a post is labeled as {\sc
  factual} or {\sc feeling} if it matches at least three high-precision
patterns for that category. If a document contains three patterns for
both categories, then we leave it unlabeled.  We ran the bootstrapping
algorithm for four iterations. 

\begin{table}[b]
\caption{Evaluation Results}
\begin{center}
\begin{tabular}{| c | c | c | c | c |}
 \hline
& \multicolumn{2}{c}{Fact}& \multicolumn{2}{|c|}{Feel} \\ \hline
& Prec & Rec & Prec & Rec \\ \hline
 \multicolumn{5}{|c|}{\cellcolor[gray]{0.9}Pattern-based Classification} \\ \hline
 Iter 0 & 77.5 & 22.8 & 65.5 & 8.0 \\
 Iter 1 & 80.0 & 34.6 & 60.0 & 16.3 \\
 Iter 2 & 80.0 & 38.0 & 64.3 & 18.8 \\
 Iter 3 & 79.9 & 40.1 & 63.0 & 19.2 \\
 Iter 4 & 78.0 & 40.9 & 62.5 & 18.8 \\ \hline
 \multicolumn{5}{|c|}{\cellcolor[gray]{0.9}Naive Bayes Classifier}
\\ \hline
 NB & 73.0 & 67.0 & 57.0 & 65.0 \\ \hline
 \end{tabular}
\end{center}
\label{table:boot_results}
\end{table}

The upper section of Table \ref{table:boot_results} shows the
Precision and Recall results for the patterns
learned during bootstrapping. The Iter 0 row shows the
performance of the patterns learned only from the original, annotated
training data. The remaining rows show the results for the patterns
learned from the unannotated texts during bootstrapping, added
cumulatively. We show the results after each iteration of
bootstrapping. 


Table \ref{table:boot_results} shows that  recall increases after each
bootstrapping iteration, demonstrating that the patterns learned from
the unannotated texts yield substantial gains in coverage over those
learned only from the annotated texts. 
Recall increases from 22.8\% to 40.9\% for {\sc fact}, and
from 8.0\% to 18.8\% for {\sc feel}.\footnote{The decrease from 19.2\%
  to 18.8\% recall is probably due to more posts being labeled as relevant by \textit{both} categories, in which case they
  are ultimately left unlabeled to avoid overlap.}
The precision for the {\sc factual} class is reasonably good, but the
precision for the {\sc feeling} class is only moderate. 
However, although precision typically decreases during
boostrapping due to the addition of imperfectly labeled data, 
the precision drop during bootstrapping is relatively small.

\begin{table*}[tb]
\begin{center}
\begin{small}
\caption{Examples of Characteristic Argumentation Style Patterns for Each Class}
\begin{tabular}{| c |  c | c | c | c |}
  \hline
Patt ID\# & Probability & Frequency & Pattern & Text Match \\ \hline
  \multicolumn{5}{|c|}{   \cellcolor[gray]{0.9}FACT Selected Patterns} \\ \hline
\bf FC1 &  1.00 & 18 & NP Prep $<$np$>$ & \texttt{SPECIES OF} \\
\bf FC2 & 1.00 & 21 & $<$subj$>$  PassVP & \texttt{EXPLANATION OF} \\ 
\bf FC3 & 1.00 & 20 & $<$subj$>$   AuxVP  Dobj & \texttt{BE EVIDENCE} \\
\bf FC4 & 1.00 & 14 & $<$subj$>$  PassVP & \texttt{OBSERVED} \\ 
\bf FC5 & 0.97 & 39 & NP Prep $<$np$>$ & \texttt{RESULT OF} \\
\bf FC6 & 0.90 & 10 & $<$subj$>$   ActVP  Dobj & \texttt{MAKE POINT} \\ 
\bf FC7 & 0.84 & 32 & Adj   Noun & \texttt{SCIENTIFIC THEORY} \\
\bf FC8 & 0.75 & 4 & NP Prep $<$np$>$ & \texttt{MISUNDERSTANDING OF} \\
\bf FC9 & 0.67 & 3 & Adj   Noun & \texttt{FUNDAMENTAL RIGHTS} \\
\bf FC10 &  0.50 & 2 & NP   Prep   $<$np$>$ & \texttt{MEASURABLE AMOUNT} \\ 
  \hline
  \multicolumn{5}{|c|}{   \cellcolor[gray]{0.9}FEEL Selected Patterns} \\ \hline
\bf FE1 &  1.00 & 14 & Adj   Noun & \texttt{MY ARGUMENT} \\   
\bf FE2 &  1.00 & 7 & $<$subj$>$   AuxVP   Adjp & \texttt{BE ABSURD} \\
\bf FE3 &  1.00 & 9 & Adv   Adj & \texttt{MORALLY WRONG} \\ 
\bf FE4 &  0.91 & 11 & $<$subj$>$   AuxVP   Adjp & \texttt{BE SAD} \\
\bf FE5 &  0.89 & 9 & $<$subj$>$   AuxVP   Adjp & \texttt{BE DUMB} \\ 
\bf FE6 &  0.89 & 9 & Adj   Noun & \texttt{NO BRAIN} \\ 
\bf FE7 &  0.81 & 37 & Adj   Noun & \texttt{COMMON SENSE} \\ 
\bf FE8 &  0.75 & 8 & InfVP   Prep   $<$np$>$ & \texttt{BELIEVE IN} \\ 
\bf FE9 &  0.87 & 3 & Adj  Noun & \texttt{ANY CREDIBILITY} \\ 
\bf FE10 &  0.53 & 17 & Adj  Noun & \texttt{YOUR OPINION} \\ 
  \hline
  \end{tabular} 
\label{table:highpatterns}
\end{small}
\end{center}
\end{table*}

We also evaluated the performance of a Naive Bayes (NB) classifier to assess the
difficulty of this task with a traditional supervised learning algorithm. We
trained a Naive Bayes classifier with unigram features and binary
values on the training data, and identified the best Laplace smoothing parameter using
the development data. The bottom row of Table \ref{table:boot_results}
shows the results for the NB classifier on the test data. 
These results show that  the NB classifier yields substantially higher
recall for both categories, undoubtedly due to the fact that the classifier uses
all unigram information available in the text.
Our pattern learner, however, was restricted to
learning linguistic expressions in specific syntactic constructions,
usually requiring more than one word, because our goal was to
study \textit{specific} expressions  associated with {\sc factual} and {\sc
  feeling} argument styles. 
Table \ref{table:boot_results} shows that the 
lexico-syntactic patterns did obtain higher precision than the NB
classifier, but with lower recall.

\begin{table}[htb]
\caption{Number of New Patterns Added after Each Round of Bootstrapping}
\centering
\begin{tabular}{| c | c | c | c |}
\hline
 & FACT & FEEL & Total\\ \hline
Iter 0 & 1,212 & 662 & 1,874\\ \hline
Iter 1 & 2,170 & 1,609 & 3,779\\
Iter 2 & 2,522 & 1,728 & 4,520\\
Iter 3 & 3,147 & 2,037 & 5,184 \\
Iter 4 & 3,696 & 2,134 & 5,830\\
\hline
\end{tabular} 
\label{table:caseframe_counts}
\end{table}

Table~\ref{table:caseframe_counts} shows the number of
patterns learned from the annotated data (Iter 0) and the number of 
new patterns added after each bootstrapping iteration. The
first iteration dramatically increases the set of
patterns, and more patterns are steadily added
throughout the rest of bootstrapping process. 


The key take-away from this set of experiments is that
distinguishing {\sc factual} and {\sc feeling} argumets is clearly a
challenging task. There is substantial room for improvement for both
precision and recall, and surprisingly, the {\sc feeling} class seems
to be harder to accurately recognize than the {\sc factual} class. In the next
section, we examine the learned patterns and their syntactic forms 
to better understand the language used in the debate forums.



\section{Analysis}
\label{analysis-sec}

\begin{figure*}
        \centering
        \begin{subfigure}[b]{0.5\textwidth}
                \includegraphics[width=\textwidth]{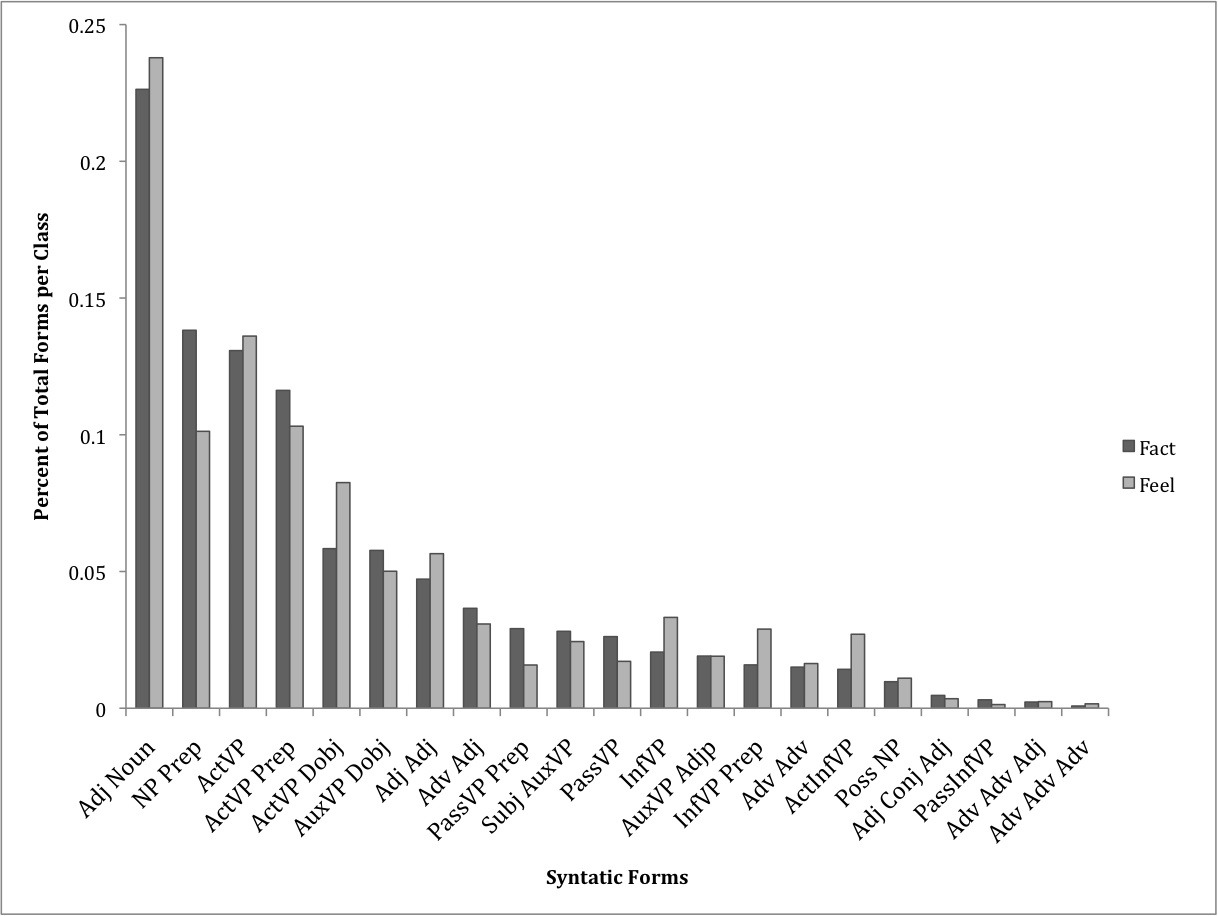}
                \caption{Percentage of Each Unique Syntactic Form}
                \label{fig:hist_unique}
        \end{subfigure}%
        ~ 
        \begin{subfigure}[b]{0.5\textwidth}
                \includegraphics[width=\textwidth]{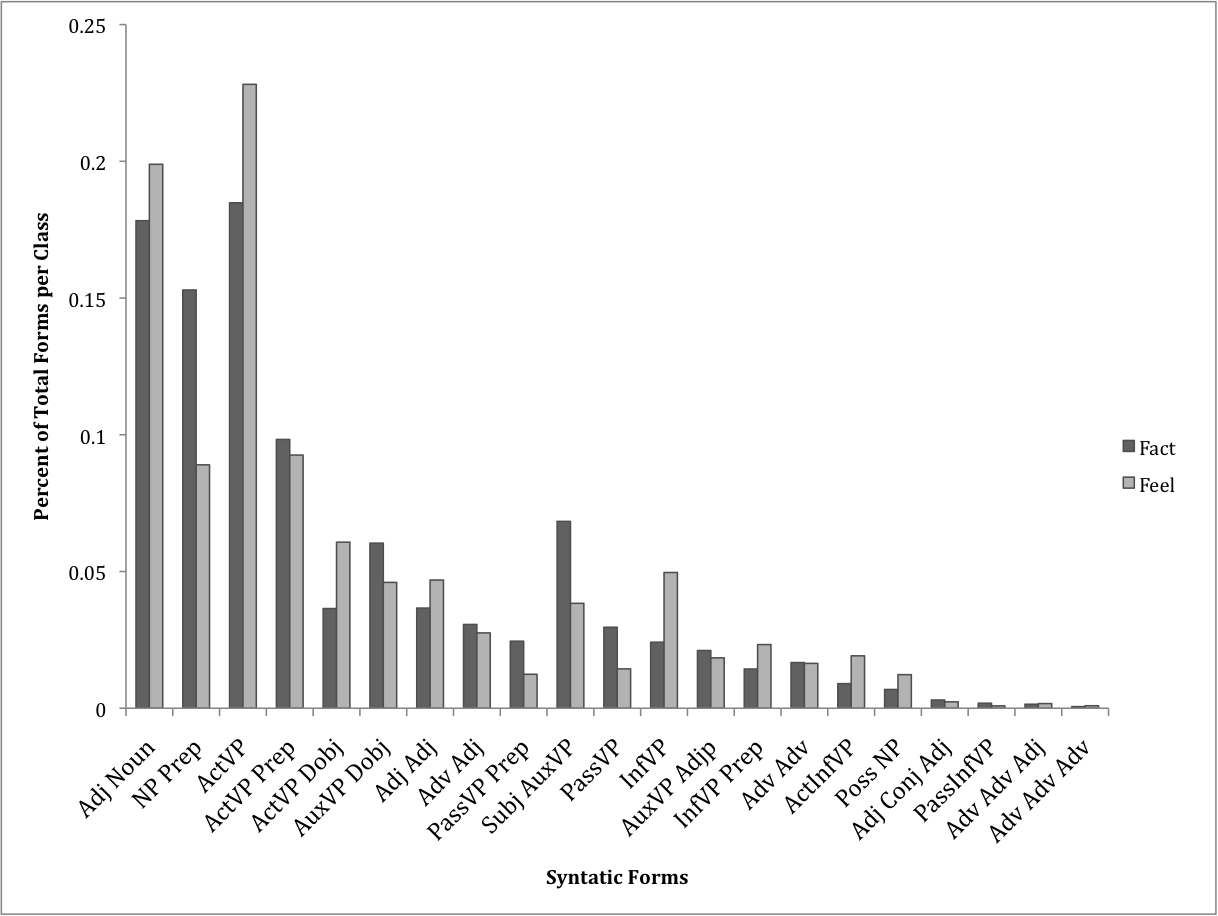}
                \caption{Percentage of Each Syntactic Form, by Instance Counts}
                \label{fig:hist_instance}
        \end{subfigure}
        \caption{Histograms of Syntactic Forms by Percentage of Total}\label{fig:hist_syntacticforms}
\end{figure*}


Table \ref{table:highpatterns} provides examples of patterns learned
for each class that are characteristic of that class.  We observe that
patterns associated with factual arguments often include
topic-specific terminology, explanatory language, and argument
phrases. In contrast, the patterns associated with feeling based
arguments are often based on the speaker's own beliefs or claims,
perhaps assuming that they themselves are credible
\cite{Chaiken80,Pettyetal81}, or they involve assessment or
evaluations of the arguments of the other speaker
\cite{hassanetal10}. They are typically also very creative and
diverse, which may be why it is hard to get higher accuracies for
{\sc feeling} classification, as shown by
Table~\ref{table:boot_results}.


Figure \ref{fig:hist_syntacticforms} shows the distribution of
syntactic forms (templates) among all of the high-precision patterns
identified for each class during bootstrapping. The x-axes show the
syntactic templates\footnote{We grouped a few of the comparable
  syntactic forms together for the purposes of this graph.} and the
y-axes show the percentage of all patterns that had a specific
syntactic form.  Figure \ref{fig:hist_unique} counts each
lexico-syntactic pattern only once, regardless of how many times it
occurred in the data set.  Figure \ref{fig:hist_instance} counts the
number of instances of each lexico-syntactic pattern.  For example,
Figure \ref{fig:hist_unique} shows that the \textit{Adj Noun}
syntactic form produced 1,400 different patterns, which comprise
22.6\% of the distinct patterns learned.  Figure
\ref{fig:hist_instance} captures the fact that there are 7,170
instances of the \textit{Adj Noun} patterns, which comprise 17.8\% of
all patterns instances in the data set. 


For {\sc factual} arguments, we see that patterns with prepositional
phrases (especially {\it NP Prep}) and passive voice verb phrases are
more common.  Instantiations of {\it NP Prep} are illustrated by {\bf
  FC1}, {\bf FC5}, {\bf FC8}, {\bf FC10} in
Table~\ref{table:highpatterns}.  Instantiations of {\it PassVP} are
illustrated by {\bf FC2} and {\bf FC4} in
Table~\ref{table:highpatterns}.  For {\sc feeling} arguments,
expressions with adjectives and active voice verb phrases are more
common. Almost every high probability pattern for {\sc feeling}
includes an adjective, as illustrated by every pattern {\bf except
  FE8} in Table~\ref{table:highpatterns}.  Figure
\ref{fig:hist_instance} shows that three syntactic forms account for a
large proportion of the instances of high-precision patterns in the
data: {\it Adj Noun}, {\it NP Prep}, and {\it ActVP}.

\begin{figure}[b]
 \centering
  \includegraphics[width=\columnwidth]{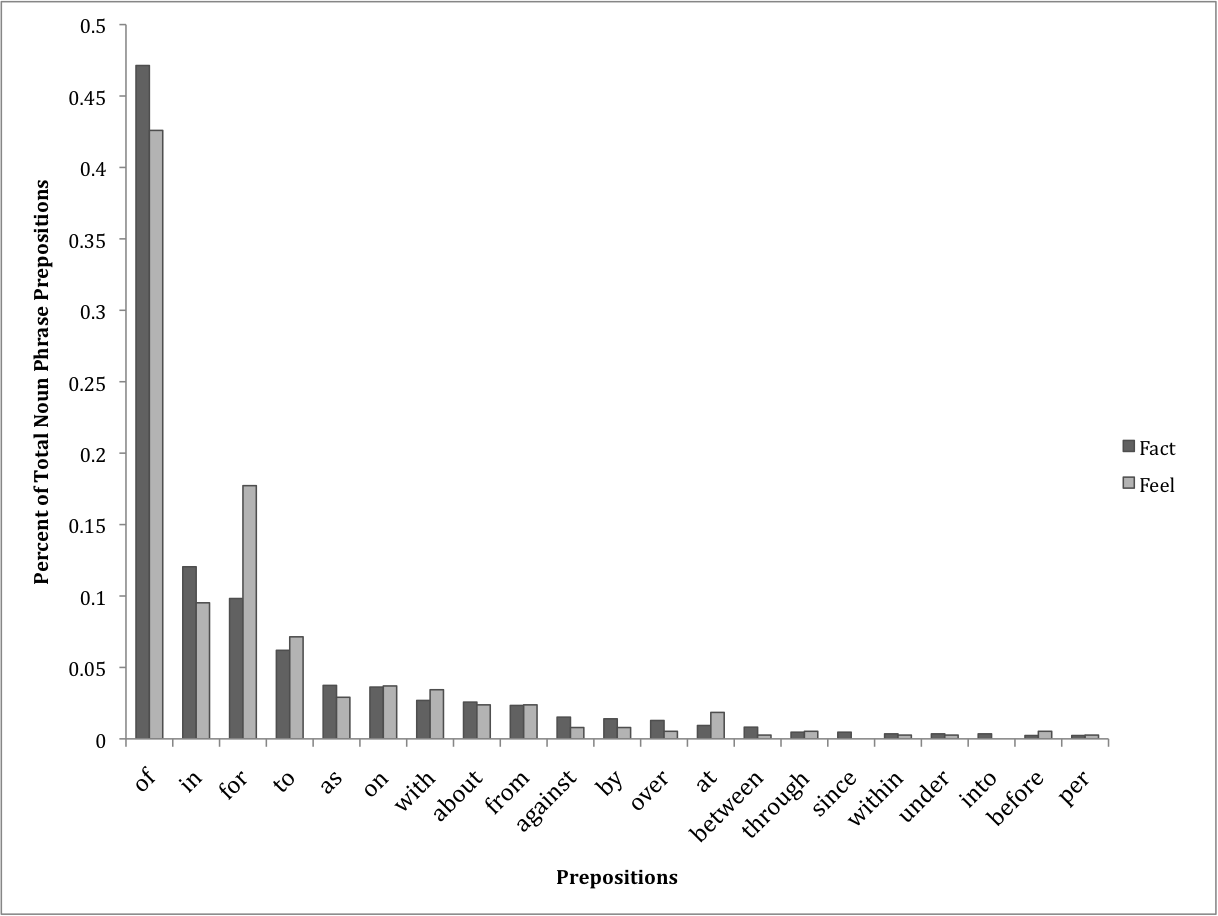}
 \caption{Percentage of Preposition Types in the {\it NP Prep} Patterns}
 \label{fig:hist_prep}
\end{figure}

Next, we further examine the {\it NP Prep} patterns since they are so
prevalent. Figure \ref{fig:hist_prep} shows the percentages of the
most frequently occurring prepositions found in the {\it NP Prep}
patterns learned for each class. Patterns containing the preposition
"of" make up the vast majority of prepositional phrases for both the
{\sc fact} and {\sc feel} classes, but is more common in the {\sc
  fact} class.  In contrast, we observe that patterns with the
preposition ``for'' are substantially more common in the {\sc feel}
class than the {\sc fact} class.

Table \ref{table:highprobpreps} shows examples of learned {\it NP
  Prep} patterns with the preposition "of" in the {\sc fact} class and
"for" in the {\sc feel} class. 
The "of" preposition in the factual arguments often attaches to 
objective terminology. 
The  "for" preposition in the feeling-based arguments is commonly used
to express advocacy (e.g., {\it demand for}) or refer to affected
population groups (e.g., {\it treatment for}). Interestingly, these phrases are subtle indicators 
of feeling-based arguments rather than explicit expressions of emotion or sentiment.

\begin{table}[t!bh]
\centering
\caption{High-Probability FACT Phrases with "OF" and FEEL Phrases with "FOR"}
\begin{small}
\begin{tabular}{| c | c |}
\hline
FACT "OF" Phrases & FEEL "FOR" Phrases \\ \hline
\texttt{RESULT OF}     & \texttt{MARRIAGE FOR}         \\
\texttt{ORIGIN OF}     & \texttt{STANDING FOR}         \\
\texttt{THEORY OF}     & \texttt{SAME FOR}             \\
\texttt{EVIDENCE OF}   & \texttt{TREATMENT FOR}        \\
\texttt{PARTS OF}      & \texttt{DEMAND FOR}           \\
\texttt{EVOLUTION OF}  & \texttt{ATTENTION FOR}        \\
\texttt{PERCENT OF}    & \texttt{ADVOCATE FOR}         \\
\texttt{THOUSANDS OF}  & \texttt{NO EVIDENCE FOR} \\
\texttt{EXAMPLE OF}    & \texttt{JUSTIFICATION FOR}    \\
\texttt{LAW OF}        & \texttt{EXCUSE FOR}           \\
\hline
\end{tabular}
\label{table:highprobpreps}
\end{small}
\end{table}

\normalsize
\section{Related Work}
\label{related-sec}

Related research on argumentation has primarily worked with different
genres of argument than found in IAC, such as news articles, weblogs,
legal briefs, supreme court summaries, and congressional debates
\cite{Marwell67,Thomasetal06,Burfoot08,Cialdini00,mcalisteretal14,ReedRowe04}. The
examples from IAC in Figure~\ref{sample-fact-feeling} illustrate that
natural informal dialogues such as those found in online forums
exhibit a much broader range of argumentative styles.  Other work has
on models of natural informal arguments 
have focused on stance classification
\cite{SomasundaranWiebe09,SomasundaranWiebe10,Walkeretal12c}, 
argument summarization \cite{Misraetal15}, sarcasm detection
\cite{Justoetal14}, and identifying the structure of arguments
such as main claims and their justifications
\cite{BiranRambow11,Purpuraetal08,YangCardie13}.

Other types of language data also typically contains a mixture of
subjective and objective sentences, e.g. Wiebe et al.
\shortcite{wiebeetal2001a,wiebeetalcl04} found that 44\% of sentences
in a news corpus were subjective.  Our work is also related to
research on distinguishing subjective and objective text
\cite{yuvh03,riloff-aaai05,wiebe05}, including bootstrapped pattern
learning for subjective/objective sentence classification
\cite{riloff-emnlp03}. However, prior work has primarily focused on
news texts, not argumentation, and the notion of objective language is
not exactly the same as factual.  Our work also aims to recognize
emotional language specifically, rather than all forms of subjective
language. There has been substantial work on sentiment and opinion
analysis (e.g.,
\cite{lee-emnlp02,kimhovy04,wilson-emnlp05,bethard05,wilson06,yang-acl14})
and recognition of specific emotions in text
\cite{mohammad2012a,mohammad2012b,Roberts12,QadirRiloff13}, which
could be incorporated in future extensions of our work. We also hope
to examine more closely the relationship of this work to previous
work aimed at the identification of nasty vs. nice
arguments in the IAC \cite{LukinWalker13,Justoetal14}.

\section{Conclusion}
\label{conc-sec}
In this paper, we use observed differences in argumentation styles in
online debate forums to extract patterns that are highly correlated
with factual and emotional argumentation. From an annotated set of
forum post responses, we are able extract high-precision patterns that
are associated with the argumentation style classes, and we are then
able to use these patterns to get a larger set of indicative patterns
using a bootstrapping methodology on a set of unannotated posts.

From the learned patterns, we derive some characteristic syntactic
forms associated with the {\sc fact} and {\sc feel} that we use to
discriminate between the classes. We observe distinctions between the
way that different arguments are expressed, with respect to the
technical and more opinionated terminologies used, which we analyze on
the basis of grammatical forms and more direct syntactic patterns,
such as the use of different prepositional phrases. Overall, we
demonstrate how the learned patterns can be used to more precisely
gather similarly-styled argument responses from a pool of unannotated
responses, carrying the characteristics of factual and emotional
argumentation style.

In future work we aim to use these insights about argument structure
to produce higher performing classifiers for identifying {\sc factual}
vs. {\sc feeling} argument styles. We also hope to understand in more
detail the relationship between these argument styles and the 
heurstic routes to persuasion and associated strategies that have been
identified in previous work on argumentation and
persuasion \cite{Marwell67,Cialdini00,ReedRowe04}.

\section*{Acknowledgments}
This work was funded by NSF Grant IIS-1302668-002 under the Robust
Intelligence Program.  The collection and annotation of the IAC corpus
was supported by an award from NPS-BAA-03 to UCSC and an IARPA Grant
under the Social Constructs in Language Program to UCSC by subcontract
from the University of Maryland.

\bibliographystyle{naaclhlt2015}
\bibliography{nl}

\end{document}